\pdfoutput=1

\documentclass[11pt]{article}

\usepackage[]{acl}

\usepackage{times}
\usepackage{latexsym}

\usepackage[T1]{fontenc}

\usepackage[utf8]{inputenc}

\usepackage{microtype}

\usepackage{inconsolata}

\usepackage{graphicx,xcolor}
\usepackage{url}

\usepackage{hyperref}

\usepackage{amsmath}
\usepackage{booktabs}
\usepackage{multirow}
\usepackage{multicol}
\usepackage{stfloats}
\usepackage{framed}
\usepackage{xurl}
\usepackage{here}
\usepackage{natbib}
\usepackage{colortbl}
\usepackage{xcolor}

\title{Are Social Sentiments Inherent in LLMs?\\
An Empirical Study on Extraction of Inter-demographic Sentiments
}

\author{
  Kunitomo Tanaka \hspace{2em} Ryohei Sasano \hspace{2em} Koichi Takeda \\
  Graduate School of Informatics, Nagoya University, Japan\\
  \texttt{tanaka.kunitomo.z3@s.mail.nagoya-u.ac.jp}, \\
  \texttt{\{sasano,takedasu\}@i.nagoya-u.ac.jp} \\
}

\begin{document}
\maketitle

\begin{abstract}
Large language models (LLMs) are supposed to acquire unconscious human knowledge and feelings, such as social common sense and biases, by training models from large amounts of text.
However, it is not clear how much the sentiments of specific social groups can be captured in various LLMs.
In this study, we focus on social groups defined in terms of nationality, religion, and race/ethnicity, and validate the extent to which sentiments between social groups can be captured in and extracted from LLMs. 
Specifically, we input questions regarding sentiments from one group to another into LLMs, apply sentiment analysis to the responses, and compare the results with social surveys.
The validation results using five representative LLMs showed higher correlations with relatively small p-values for nationalities and religions, whose number of data points were relatively large. This result indicates that the LLM responses including the inter-group sentiments align well with actual social survey results.
\end{abstract}

\section{Introduction}
Large language models (LLMs) can generate high-quality text indistinguishable from human-generated text for a variety of tasks~\cite{gpt4,llama2}.
Accordingly, several attempts have been made to reproduce social experiments with LLMs instead of surveys with human subjects, focusing on their ability to imitate human behavior and dialog~\cite{employing, human-subject, homo-silicus, multi-agent}.

\begin{figure}
    \centering
    \includegraphics[width=\columnwidth]{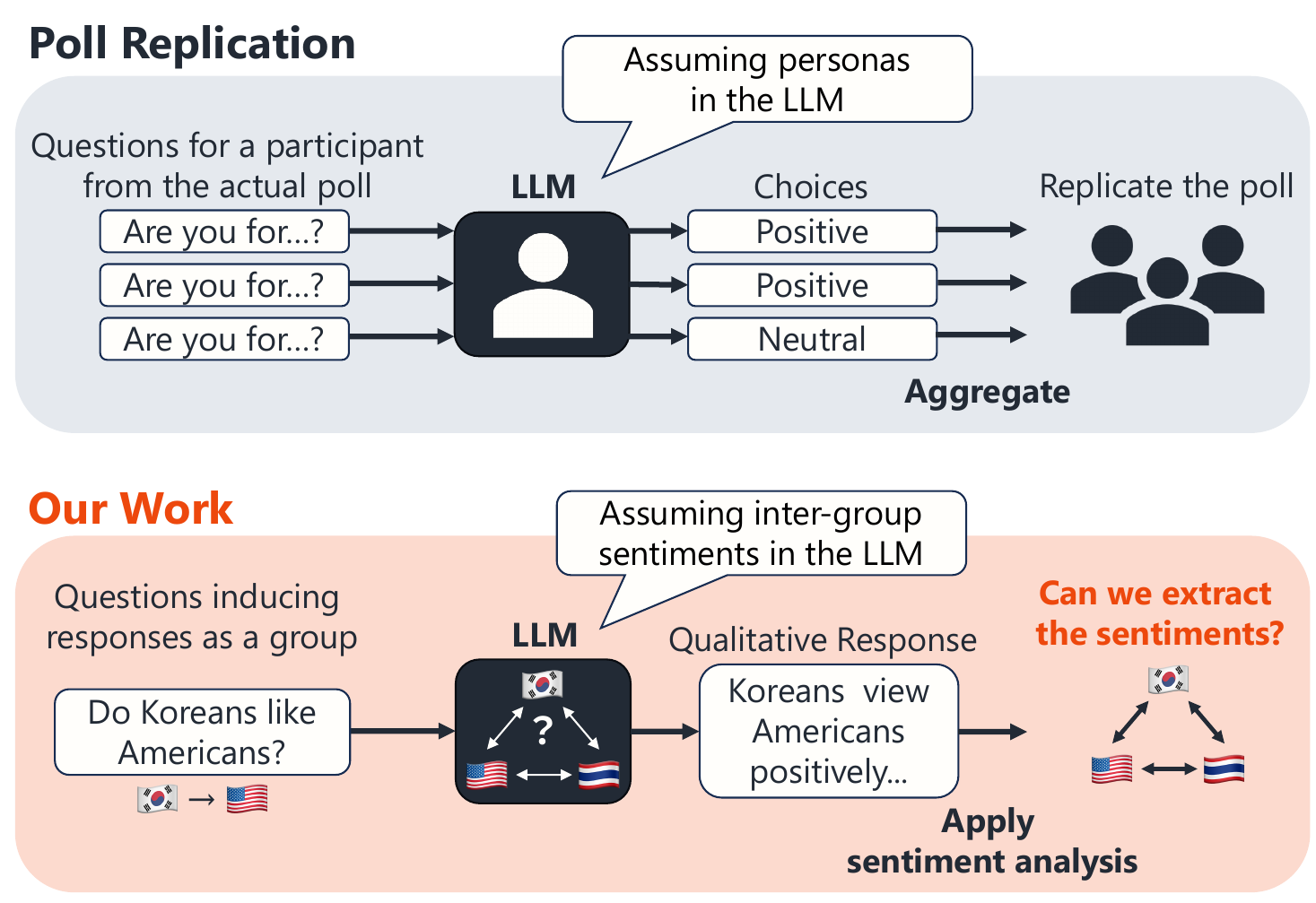}
    \caption{Motivation of our work contrasted with prior works~\cite{out-of-one, whose-opinion, global, random}.}
    \label{fig:motivation}
\end{figure}

Among the studies that employ LLMs as a substitute for humans, there is a growing trend of reproducing opinion polls~\cite{out-of-one,whose-opinion,global,questioning,random} as the survey cost escalates commensurately with their scale.
Most of these studies make an effort to reproduce the collective conceptions on social issues by instructing LLMs to respond to the input of the questions and the corresponding choices from the actual polls. 
In contrast, it remains unclear to what extent sentiments held by specific social groups can be extracted from various LLMs.
Thus, we rather hypothesize that the LLMs potentially harbor knowledge and sentiments as each whole demographic group, and focus on how much of inter-group sentiment can be extracted from their outputs, as outlined in Figure~\ref{fig:motivation}.
Specifically, we consider three attributes that define social groups: nationalities, religions, and races/ethnicities.

\section{Related Work} \label{sec:related}
The rapid development of LLMs has been accompanied by an increasing number of studies that have substituted outputs from LLMs for human responses, demonstrating that they can replicate human behaviors in psychology~\cite{human-subject}, economics~\cite{homo-silicus} and many others simulating multi-agents~\cite{multi-agent}.
As such, LLMs might reproduce complex words and actions by humans and could well encompass human sentiments; hence the emergence of attempts to reproduce the results of opinion polls using LLMs~\cite{employing}.
Many of them input the questions and the choices from actual opinion polls to replicate it while prompting them to imitate demographic personas.
\citet{out-of-one} have shown the potential that LLMs replicate group-specific trends by giving them personas of social survey participants, such as age and gender, and then having them answer a social survey on U.S. politics. 
The study by \citet{random} has advanced the feasibility of the approach by \citet{out-of-one}. Their findings indicate that the method effectively replicates opinions to a significant extent. Nonetheless, differences in how well the model replicates results for various demographic groups reveal an underlying bias in the language model.
Also, \citet{whose-opinion} have indicated that the LLM is less likely to reflect opinions in the U.S. especially for minority views even given persona although the LLMs are tuned aligned with human preference.
As for the opinion replication on a global scale, \citet{global} have pointed out that LLMs are biased towards Western values, which means LLMs may not necessarily replicate the poll results for participants of the target nationality.
Moreover, some studies have indicated that multiple-choice questions might not be suitable for the reproduction with LLMs~\cite{arrow}.

One reason for unfair tendencies of LLMs is that, while learning knowledge and sentiments from a large corpus, they also internalize potential social biases present in the dataset~\cite{stochastic-parrot, bias-in-nlp}.
For example, language models are known to learn a broad spectrum of biases, including those related to gender~\cite{out-of-box, gender}, nationality~\cite{nationality}, religion~\cite{anti-muslim}, and race~\cite{survey-race}.
LLMs are therefore susceptible to the demographics assigned to them, resulting in the skew of the outputs~\cite{incontext-impersonation, bias-runs-deep, cannot}.
Despite the growing trend of the studies on survey replication, the extent to which how much LLM's responses express collective social sentiment is yet to be examined to the best of our knowledge.
In this paper, we rather focus on the replication assuming LLMs to bear knowledge and social sentiments among demographics and validates how much of the sentiments can be extracted from open-ended responses.

\section{Target Social Groups and Data} \label{sec:data}
We consider three attributes that define demographic groups: nationalities, religions, and races/ \hspace{-0.7ex}ethnicities.
For each pair of attributes, the sentiment from group $G_\text{from}$ towards group $G_\text{to}$ is extracted from LLMs. 
In order to assess how well inter-group sentiments are extracted, we then calculate the correlation coefficient between the data from actual poll results and the scores of extracted sentiments.
Table \ref{tab:target_group} lists the social groups considered in this study for each attribute.
Below we briefly describe the data of the actual poll results.\footnote{More detailed descriptions are provided in Appendix \ref{Sec::detail-of-data}.}

\paragraph{Nationalities}
We draw the data from the polling report by the Japan Press Research Institute taken in 2022.\footnote{\url{https://www.chosakai.gr.jp/wp/wp-content/themes/shinbun/asset/pdf/project/notification/kaigaiyoron2022hodo_2.pdf\#page=11}}
The participants were given four options, and the data represents each percentage of the participants who gave the positive options of all the participants.
The table on the right in Figure \ref{fig:gpt4-nationality} illustrates the actual poll data.

\paragraph{Religions} 
We draw the data from the polling report by Pew Research Center taken in 2022~\cite{pew2023jews}. 
The participants were given six options, and the data represents each percentage of the participants who gave the positive options minus the percentage of the participants who gave the negative options.
The table on the right in Figure \ref{fig:gpt4-religion} illustrates the actual poll data.

\paragraph{Races/ethnicities} 
We draw the data from the polling report by Pew Research Center taken in 2019~\cite{pew2019views}. 
The participants were asked to score their sentiments toward another group of race/ethnicity on a scale of 0--100.
The data represents the mean score of each inter-group sentiment.
The table on the right in Figure \ref{fig:gpt4-race} illustrates the actual poll data.

\begin{table}[t]
    \centering
    \small
    \begin{tabular}{cl}
        \toprule
        Attribute  & Social Groups\\\midrule
        Nationalities & Chinese (\texttt{CN}), French (\texttt{FR}), \\
        & British (\texttt{GB}), Korean (\texttt{KR}), \\
        & Thai (\texttt{TH}), American (\texttt{US}), \\
        & Japanese* (\texttt{JP}), Russian* (\texttt{RU})\\ \midrule
        Religions & atheist (\texttt{ATH}), Catholic (\texttt{CTH}), \\
        & Evangelical (\texttt{EVG}), Jew (\texttt{JEW}), \\
        & Mainline Protestant, (\texttt{MPR}), \\
        & Mormon (\texttt{LDS}), Muslim* (\texttt{MUS})\\\midrule
        Races/ethnicities & Asian (\texttt{AS}), Black (\texttt{BL}), \\
        & Hispanic (\texttt{SP}), White (\texttt{WH})\\
        \bottomrule
    \end{tabular}
    \vspace{-0.3ex}
    \caption{List of social groups. 
    `*' indicates that the group is considered only as $G_\textrm{to}$.}
    \label{tab:target_group}
    \vspace{-0.7ex}
\end{table}

\section{Sentiment Extraction between Social Groups and their Validation} \label{chap:method}
\begin{figure*}[t]
    \centering
    \vspace{0ex}
    \includegraphics[width=\linewidth]{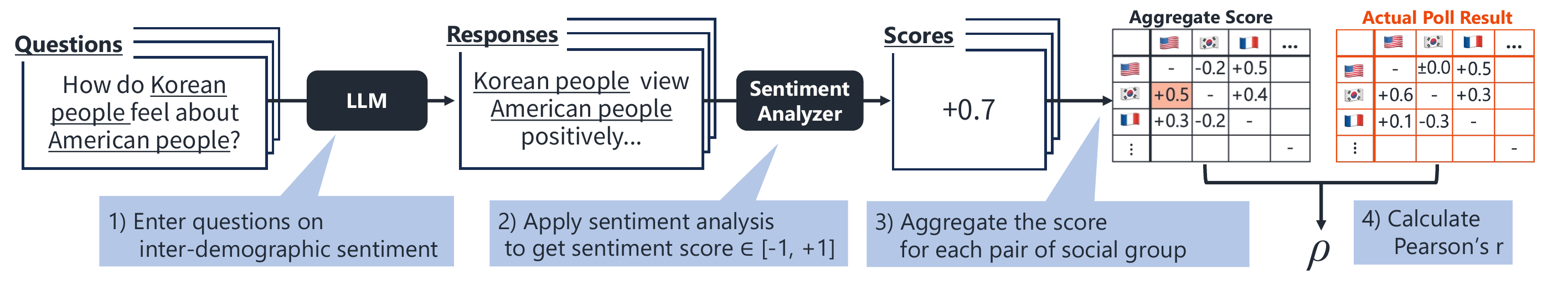}
    \vspace{-4ex}
    \caption{Procedure for extraction of sentiments between social groups from LLMs.}
    \label{fig:steps}
\end{figure*}

In this study, we validate the extent to which sentiments between social groups can be extracted from LLMs. Figure~\ref{fig:steps} shows the validation procedure that we employ.

\subsection{Questions on Inter-Demographic Sentiments}
\label{Sec::Quesion}
We prepare questions on inter-group sentiments to input to LLMs by using question templates shown in Table \ref{tab:question_template}.
We initially direct the LLMs with the system prompt ``\texttt{Always answer in English.}'', then input a question prepared with the template.
The templates are broadly divided into two types: templates for yes-no questions and templates for wh-questions, and each type is composed of six templates.
We investigate the relationship between question type and sentiment extraction performance by conducting experiments in three different settings: using only yes-no-type templates, only wh-type templates, and a mixture of types, taking every other template in Table \ref{tab:question_template} from the top.

For each template, the subject of the sentiment is assigned to $G_\text{from}$, and the object to $G_\text{to}$. 
To explicitly indicate the questions are about social groups, we slightly modify the entries in Table \ref{tab:target_group} when instantiating the templates into input to LLMs.
Specifically, we suffix the word ``people'' to every entry of nationalities or races/ethnicities, and pluralize every entry of religions. 
In addition to those modifications, the phrase ``In the U.S., '' is added before the questions on religions or races/ethnicities, for the polls of religions and races/ethnicities were taken in the U.S.
For instance, as to the sentiment from Asian people towards Black people, the template ``\texttt{Do $G_\textrm{from}$ like $G_\textrm{to}$?}'' generates the question ``\texttt{In the U.S., do Asian people like Black people?}''.

To mitigate the potential impact of randomness in the LLM responses, each question generated with the template is entered independently three times, yielding three responses.
As there are six templates in a question type, we obtain 18 responses for each question type from the LLM.

\begin{table}[t]
    \centering
    \small 
    \begin{tabular}{cl}
        \toprule
        Type &  Template  \\
        \midrule
        \multirow{6}{*}{Yes-no} & \textrm{Do $G_\textrm{from}$ have good feelings on $G_\textrm{to}$?} \\
                                & \textrm{Do $G_\textrm{from}$ have favorable feelings on $G_\textrm{to}$?} \\
                                & \textrm{Do $G_\textrm{from}$ have positive feelings on $G_\textrm{to}$?} \\
                                & \textrm{Do $G_\textrm{from}$ like $G_\textrm{to}$?} \\
                                & \textrm{Do $G_\textrm{from}$ love $G_\textrm{to}$?} \\
                                & \textrm{Do $G_\textrm{from}$ appreciate $G_\textrm{to}$?} \\
        \midrule
        \multirow{6}{*}{Wh}     & \textrm{What are $G_\textrm{from}$'s feelings on $G_\textrm{to}$?} \\
                                & \textrm{What are $G_\textrm{from}$'s impressions on $G_\textrm{to}$?} \\
                                & \textrm{What are $G_\textrm{from}$'s thoughts on $G_\textrm{to}$?} \\
                                & \textrm{How do $G_\textrm{from}$ feel about $G_\textrm{to}$?} \\
                                & \textrm{How do $G_\textrm{from}$ view $G_\textrm{to}$?} \\
                                & \textrm{How do $G_\textrm{from}$ perceive $G_\textrm{to}$?} \\
                                
        \bottomrule
    \end{tabular}
    \caption{Question templates passed to LLMs.}
    \label{tab:question_template}
\end{table}

\subsection{Score Computation of LLM Responses and Aggregation}
\label{Sec::Scoring}
We apply sentiment analysis to each response from LLMs to score the sentiments. 
Specifically, each response from LLMs is fed to a sentiment analyzer that can score the input from -1 to +1.
Finally, we determine the sentiment score for each demographic pair by calculating the average of the scores from all responses on the group pair of interest.

Next, for each attribute, we aggregate inter-group sentiments of all combinations and compare them to the corresponding actual poll result.
To make allowances for the gap between the distributions of the two, we compute the agreement between them by Pearson correlation coefficient ($\rho$) rather than an absolute difference.
The coefficient value is to illustrate the extractability of inter-group sentiments of each attribute.

\section{Experiments} \label{Sec:Exp}
We investigated the extent to which inter-group sentiments about nationalities, religions, or races/ethnicities can be extracted.

\subsection{Experimental Settings}
We selected the following five LLMs for the validation. 
Default settings were used for each model.
\begin{itemize}
\setlength{\parskip}{0.3ex}
\setlength{\itemsep}{0ex}
\item GPT-3.5 Turbo (\texttt{gpt-3.5-turbo-0613}\footnote{\scriptsize \url{https://platform.openai.com/docs/models/gpt-3-5}})
\item GPT-4 (\texttt{gpt-4-preview-1106}\footnote{\scriptsize \url{https://platform.openai.com/docs/models/gpt-4}})
\item Llama 2-Chat 13B\footnote{\scriptsize \url{https://huggingface.co/meta-llama/Llama-2-13b-chat}}
\item Llama 2-Chat 70B\footnote{\scriptsize \url{https://huggingface.co/meta-llama/Llama-2-70b-chat}}
\item Vicuna 13B v1.5\footnote{\scriptsize \url{https://huggingface.co/lmsys/vicuna-13b-v1.5}}
\end{itemize}
In order to extract sentiments, we employ VADER~\cite{vader}, a sentiment analyzer calculating the psychological valence of each word in an input and outputs the score $\in [-1, +1]$.

As examples of the tables for computing correlations, Appendix \ref{Sec:Sample} provides Figures \ref{fig:gpt4-nationality}, \ref{fig:gpt4-religion}, and \ref{fig:gpt4-race}.
They are sample tables showing sentiment scores alongside the results of actual social surveys for nationality, religion, and race/ethnicity.

\subsection{Results} \label{subsec:exp-result}

\begin{figure*}[t]
    \begin{minipage}[t]{0.8\textwidth}
    \begin{flushright}
        \includegraphics[width=\linewidth]{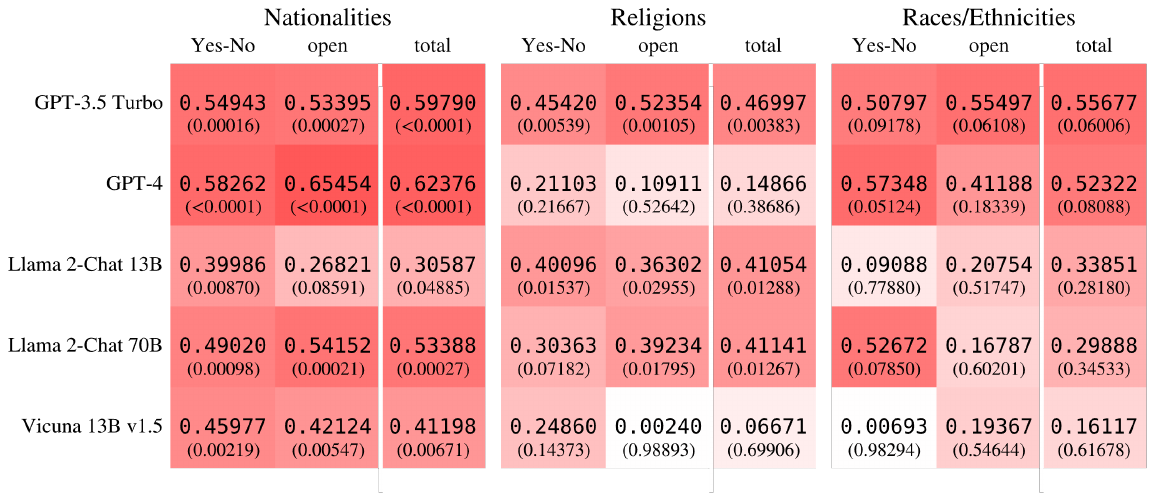}
        \end{flushright}
    \end{minipage}
  \hfill
    \begin{minipage}[t]{0.12\textwidth}
    \vspace{-6ex}
    \footnotesize
      \begin{tabular}{@{}>{\columncolor{pink}}c@{}}
       $\rho$\\[-1pt]
       \tiny{($p$-value)}\\
      \end{tabular}
    \end{minipage}
    \vspace{-0.5ex}
    \caption{Correlation coefficients between the actual poll result and the sentiment scores for each combination of LLMs and the set of question templates. Below them show the p-values of non-correlation test.}
    \label{fig:correlations-vader}
    \vspace{-1ex}
\end{figure*}

Figure \ref{fig:correlations-vader} shows the correlation coefficients $\rho$ for each combination of LLMs and the set of question templates for nationalities, religions, and races/ethnicities.
The values in parentheses under $\rho$ denote the p-values for testing non-correlation.\footnote{The p-values for testing non-correlation were calculated using \texttt{scipy.stats.pearsonr} in SciPy (\url{https://scipy.org/}).}

Overall, positive correlation coefficients were obtained in all cases, indicating that LLMs align with real inter-demographic sentiments.
For nationality data in particular, the correlation coefficients usually surpass 0.3, regardless of LLMs or sentiment classifiers, except when the Llama 2-Chat 13B responds to wh-type questions. 
For the religion data, while lower correlation coefficients were observed more frequently than for the nationality data, more than half of the combinations shows correlations of 0.3 or higher. 
Also, higher p-values were observed in many cells, suggesting that the results are more variable than those by nationality.
Similarly, high correlations were observed with race/ethnicity, but it should be noted that the p-values were high (mostly exceeding 0.1), which is presumably for scarcity of the data points (12 points).

As for the question type, we expected wh-questions to elicit open-ended answers as the LLMs would connote nuanced social sentiments, considering that the latest LLMs tend to avoid providing direct answers to sensitive questions. 
However, no evident differences were observed between yes-no-type and wh-type questions.

Figures \ref{fig:gpt4-nationality}, \ref{fig:gpt4-religion}, and \ref{fig:gpt4-race} provide sample tables showing sentiment scores alongside the results of actual social surveys for nationality, religion, and race/ethnicity in that order from the top.
Each figure shows the mean score by each group pair of the 18 responses to six yes-no questions on the left, 
and the table in the center for six wh-questions.
$\rho$ on those tables indicate the correlation coefficient with the actual poll result on the right.
The vertical axis indicates the subject of the sentiment $G_\text{from}$ and the horizontal axis indicates the object of the sentiment $G_\text{to}$.

\section{Conclusion} \label{sec:conclude}
In this paper, we have validated the extent to which LLMs express inter-demographic sentiments defined by nationalities, religions, and races/ethnicities in their qualitative responses 
by inputting questions related to sentiments between two groups into LLMs and applying sentiment analysis to their responses.
The validation results using five representative LLMs showed higher correlations with relatively small p-values for nationalities and religions, whose number of data points were relatively large. 
This result suggests that the LLM responses contain the sentiments among social groups, aligned with actual ones.
However, our experiments were only conducted on three attributes in English and thus need to be performed on more languages and social groups to draw more general conclusions, which is our future work.

\bibliography{custom}

\begin{thebibliography}{25}
\expandafter\ifx\csname natexlab\endcsname\relax\def\natexlab#1{#1}\fi

\bibitem[{Abid et~al.(2021)Abid, Farooqi, and Zou}]{anti-muslim}
Abubakar Abid, Maheen Farooqi, and James Zou. 2021.
\newblock {Persistent Anti-Muslim Bias in Large Language Models}.
\newblock In \emph{Proceedings of the 2021 AAAI/ACM Conference on AI, Ethics, and Society (AIES)}, pages 298--306.

\bibitem[{Aher et~al.(2023)Aher, Arriaga, and Kalai}]{human-subject}
Gati~V Aher, Rosa~I. Arriaga, and Adam~Tauman Kalai. 2023.
\newblock {Using Large Language Models to Simulate Multiple Humans and Replicate Human Subject Studies}.
\newblock In \emph{Proceedings of the 40th International Conference on Machine Learning (ICML)}, pages 337--371.

\bibitem[{Argyle et~al.(2023)Argyle, Busby, Fulda, Gubler, Rytting, and Wingate}]{out-of-one}
Lisa~P Argyle, Ethan~C Busby, Nancy Fulda, Joshua~R Gubler, Christopher Rytting, and David Wingate. 2023.
\newblock {Out of One, Many: Using Language Models to Simulate Human Samples}.
\newblock \emph{Political Analysis}, 31(3):337--351.

\bibitem[{Bender et~al.(2021)Bender, Gebru, McMillan-Major, and Shmitchell}]{stochastic-parrot}
Emily~M. Bender, Timnit Gebru, Angelina McMillan-Major, and Shmargaret Shmitchell. 2021.
\newblock {On the Dangers of Stochastic Parrots: Can Language Models Be Too Big?}
\newblock In \emph{Proceedings of the 2021 ACM Conference on Fairness, Accountability, and Transparency (FAccT)}, page 610–623.

\bibitem[{Blodgett et~al.(2020)Blodgett, Barocas, Daum{\'e}~III, and Wallach}]{bias-in-nlp}
Su~Lin Blodgett, Solon Barocas, Hal Daum{\'e}~III, and Hanna Wallach. 2020.
\newblock Language (technology) is power: A critical survey of {``}bias{''} in {NLP}.
\newblock In \emph{Proceedings of the 58th Annual Meeting of the Association for Computational Linguistics}, pages 5454--5476.

\bibitem[{Dominguez-Olmedo et~al.(2023)Dominguez-Olmedo, Hardt, and Mendler-D{\"u}nner}]{questioning}
Ricardo Dominguez-Olmedo, Moritz Hardt, and Celestine Mendler-D{\"u}nner. 2023.
\newblock {Questioning the Survey Responses of Large Language Models}.
\newblock \emph{arXiv preprint arXiv:2306.07951}.

\bibitem[{Durmus et~al.(2023)Durmus, Nyugen, Liao, Schiefer, Askell, Bakhtin, Chen, Hatfield-Dodds, Hernandez, Joseph et~al.}]{global}
Esin Durmus, Karina Nyugen, Thomas~I Liao, Nicholas Schiefer, Amanda Askell, Anton Bakhtin, Carol Chen, Zac Hatfield-Dodds, Danny Hernandez, Nicholas Joseph, et~al. 2023.
\newblock {Towards Measuring the Representation of Subjective Global Opinions in Language Models}.
\newblock \emph{arXiv preprint arXiv:2306.16388}.

\bibitem[{Field et~al.(2021)Field, Blodgett, Waseem, and Tsvetkov}]{survey-race}
Anjalie Field, Su~Lin Blodgett, Zeerak Waseem, and Yulia Tsvetkov. 2021.
\newblock {A Survey of Race, Racism, and Anti-Racism in NLP}.
\newblock In \emph{Proceedings of the 59th Annual Meeting of the Association for Computational Linguistics and the 11th International Joint Conference on Natural Language Processing (ACL-IJCNLP)}, pages 1905--1925.

\bibitem[{Guo et~al.(2024)Guo, Chen, Wang, Chang, Pei, Chawla, Wiest, and Zhang}]{multi-agent}
Taicheng Guo, Xiuying Chen, Yaqi Wang, Ruidi Chang, Shichao Pei, Nitesh~V Chawla, Olaf Wiest, and Xiangliang Zhang. 2024.
\newblock Large language model based multi-agents: A survey of progress and challenges.
\newblock \emph{arXiv preprint arXiv:2402.01680}.

\bibitem[{Gupta et~al.(2023)Gupta, Shrivastava, Deshpande, Kalyan, Clark, Sabharwal, and Khot}]{bias-runs-deep}
Shashank Gupta, Vaishnavi Shrivastava, Ameet Deshpande, Ashwin Kalyan, Peter Clark, Ashish Sabharwal, and Tushar Khot. 2023.
\newblock {Bias Runs Deep: Implicit Reasoning Biases in Persona-Assigned LLMs}.
\newblock \emph{arXiv preprint arXiv:2311.04892}.

\bibitem[{Horowitz et~al.(2019)Horowitz, Brown, Cox, and of~Michigan. Digital Library Platform \&~Services}]{pew2019views}
J.M. Horowitz, A.~Brown, K.~Cox, and University of~Michigan. Digital Library Platform \&~Services. 2019.
\newblock \href {https://www.pewresearch.org/social-trends/2019/04/09/race-in-america-2019/} {\emph{{Race in America 2019: Public Has Negative Views of the Country's Racial Progress; More Than Half Say Trump Has Made Race Relations Worse}}}.
\newblock Pew Research Center.

\bibitem[{Horton(2023)}]{homo-silicus}
John~J Horton. 2023.
\newblock {Large Language Models as Simulated Economic Agents: What Can We Learn from Homo Silicus?}
\newblock Technical report, National Bureau of Economic Research.

\bibitem[{Hutto and Gilbert(2014)}]{vader}
Clayton Hutto and Eric Gilbert. 2014.
\newblock {VADER: A Parsimonious Rule-Based Model for Sentiment Analysis of Social Media Text}.
\newblock In \emph{Proceedings of the International AAAI Conference on Web and Social Media}, volume~8, pages 216--225.

\bibitem[{Jansen et~al.(2023)Jansen, Jung, and Salminen}]{employing}
Bernard~J Jansen, Soon-gyo Jung, and Joni Salminen. 2023.
\newblock {Employing large language models in survey research}.
\newblock \emph{Natural Language Processing Journal}, 4:100020.

\bibitem[{Kirk et~al.(2021)Kirk, Jun, Volpin, Iqbal, Benussi, Dreyer, Shtedritski, and Asano}]{out-of-box}
Hannah~Rose Kirk, Yennie Jun, Filippo Volpin, Haider Iqbal, Elias Benussi, Frederic Dreyer, Aleksandar Shtedritski, and Yuki Asano. 2021.
\newblock {Bias Out-of-the-Box: An Empirical Analysis of Intersectional Occupational Biases in Popular Generative Language Models}.
\newblock \emph{Advances in Neural Information Processing Systems (NeurIPS)}, pages 2611--2624.

\bibitem[{Lucy and Bamman(2021)}]{gender}
Li~Lucy and David Bamman. 2021.
\newblock {Gender and Representation Bias in GPT-3 Generated Stories}.
\newblock In \emph{Proceedings of the Third Workshop on Narrative Understanding (NUSE)}, pages 48--55.

\bibitem[{Narayanan~Venkit et~al.(2023)Narayanan~Venkit, Gautam, Panchanadikar, Huang, and Wilson}]{nationality}
Pranav Narayanan~Venkit, Sanjana Gautam, Ruchi Panchanadikar, Ting-Hao Huang, and Shomir Wilson. 2023.
\newblock {Nationality Bias in Text Generation}.
\newblock In \emph{Proceedings of the 17th Conference of the European Chapter of the Association for Computational Linguistics (EACL)}, pages 116--122.

\bibitem[{OpenAI(2023)}]{gpt4}
OpenAI. 2023.
\newblock {GPT-4 Technical Report}.
\newblock Technical report.

\bibitem[{R{\"o}ttger et~al.(2024)R{\"o}ttger, Hofmann, Pyatkin, Hinck, Kirk, Sch{\"u}tze, and Hovy}]{arrow}
Paul R{\"o}ttger, Valentin Hofmann, Valentina Pyatkin, Musashi Hinck, Hannah~Rose Kirk, Hinrich Sch{\"u}tze, and Dirk Hovy. 2024.
\newblock {Political Compass or Spinning Arrow? Towards More Meaningful Evaluations for Values and Opinions in Large Language Models}.
\newblock \emph{arXiv preprint arXiv:2402.16786}.

\bibitem[{Salewski et~al.(2024)Salewski, Alaniz, Rio-Torto, Schulz, and Akata}]{incontext-impersonation}
Leonard Salewski, Stephan Alaniz, Isabel Rio-Torto, Eric Schulz, and Zeynep Akata. 2024.
\newblock In-context impersonation reveals large language models' strengths and biases.
\newblock \emph{Advances in Neural Information Processing Systems}, 36.

\bibitem[{Santurkar et~al.(2023)Santurkar, Durmus, Ladhak, Lee, Liang, and Hashimoto}]{whose-opinion}
Shibani Santurkar, Esin Durmus, Faisal Ladhak, Cinoo Lee, Percy Liang, and Tatsunori Hashimoto. 2023.
\newblock {Whose Opinions Do Language Models Reflect?}
\newblock \emph{arXiv preprint arXiv:2303.17548}.

\bibitem[{Sun et~al.(2024)Sun, Lee, Nan, Zhao, Lee, Jansen, and Kim}]{random}
Seungjong Sun, Eungu Lee, Dongyan Nan, Xiangying Zhao, Wonbyung Lee, Bernard~J Jansen, and Jang~Hyun Kim. 2024.
\newblock {Random Silicon Sampling: Simulating Human Sub-Population Opinion Using a Large Language Model Based on Group-Level Demographic Information}.
\newblock \emph{arXiv preprint arXiv:2402.18144}.

\bibitem[{Tevington(2023)}]{pew2023jews}
Patricia Tevington. 2023.
\newblock \href {https://www.pewresearch.org/religion/2023/03/15/americans-feel-more-positive-than-negative-about-jews-mainline-protestants-catholics/} {\emph{{Americans Feel More Positive Than Negative About Jews, Mainline Protestants, Catholics}}}.
\newblock Pew Research Center.

\bibitem[{Touvron et~al.(2023)Touvron, Martin, Stone, Albert, Almahairi, Babaei, Bashlykov, Batra, Bhargava, Bhosale et~al.}]{llama2}
Hugo Touvron, Louis Martin, Kevin Stone, Peter Albert, Amjad Almahairi, Yasmine Babaei, Nikolay Bashlykov, Soumya Batra, Prajjwal Bhargava, Shruti Bhosale, et~al. 2023.
\newblock {Llama 2: Open Foundation and Fine-Tuned Chat Models}.
\newblock Technical report, GenAI, Meta.

\bibitem[{Wang et~al.(2024)Wang, Morgenstern, and Dickerson}]{cannot}
Angelina Wang, Jamie Morgenstern, and John~P Dickerson. 2024.
\newblock {Large Language Models Cannot Replace Human Participants because They Cannot Portray Identity Groups}.
\newblock \emph{arXiv preprint arXiv:2402.01908}.

\end{thebibliography}

\appendix
\section{Details of the Actual Poll Data} \label{Sec::detail-of-data}
\paragraph{Nationalities}
The data is drawn from the polling report by the Japan Press Research Institute taken in 2022.\footnotemark[2]
In this poll, approximately 1,000 participants per nationality were sampled to conduct a survey on media and international sentiments.
This study draws the data collected from participants who were asked about their sentiments towards another country. 
The participants were given four options: ``Very favorable'', ``Favorable'', ``Not very favorable'', and ``Not at all favorable'', 
and the data represents each percentage of the participants who gave the positive options 
(i.e., ``Very favorable'' or ``favorable'') 
of all the participants.

\paragraph{Religions} 
The data is drawn from the polling report by Pew Research Center taken in 2022~\cite{pew2023jews}.
In this poll, 10,588 participants from the United States were sampled from the panel to conduct a survey concerning religions, and the polling results are weighted to reflect the distribution of the U.S. population.
This study draws the data collected from participants who were asked about their sentiments towards another religion.
The participants were given six options: ``Very favorable'', ``Somewhat favorable'', ``Neither favorable or unfavorable'', ``Somewhat unfavorable', ``Very unfavorable'', ``Don't know enough to say'', 
and the data represents each percentage of the participants who gave the positive options (i.e., ``Very favorable'' or ``Somewhat favorable'') minus the percentage of the participants who gave the negative options (i.e., ``Somewhat unfavorable'' or ``Very unfavorable'').

\paragraph{Races/ethnicities} 
The data is drawn from the polling report by Pew Research Center taken in 2019~\cite{pew2019views}.
In this poll, 6,637 participants from the United States were sampled from the panel to conduct a survey concerning races/ethnicities, and the polling results are weighted to reflect the distribution of the U.S. population.
This study draws the data collected from participants who were asked about their sentiments towards another race/ethnicity.
The participants were asked to score their sentiments toward another group of race/ethnicity on a scale of 0--100.
The data represents the mean score of each inter-group sentiment.

\section{Sample Tables of Sentiment Scores and Actual Social Survey Result} \label{Sec:Sample}

\begin{figure*}[t!]
    \centering
    \includegraphics[width=\linewidth]{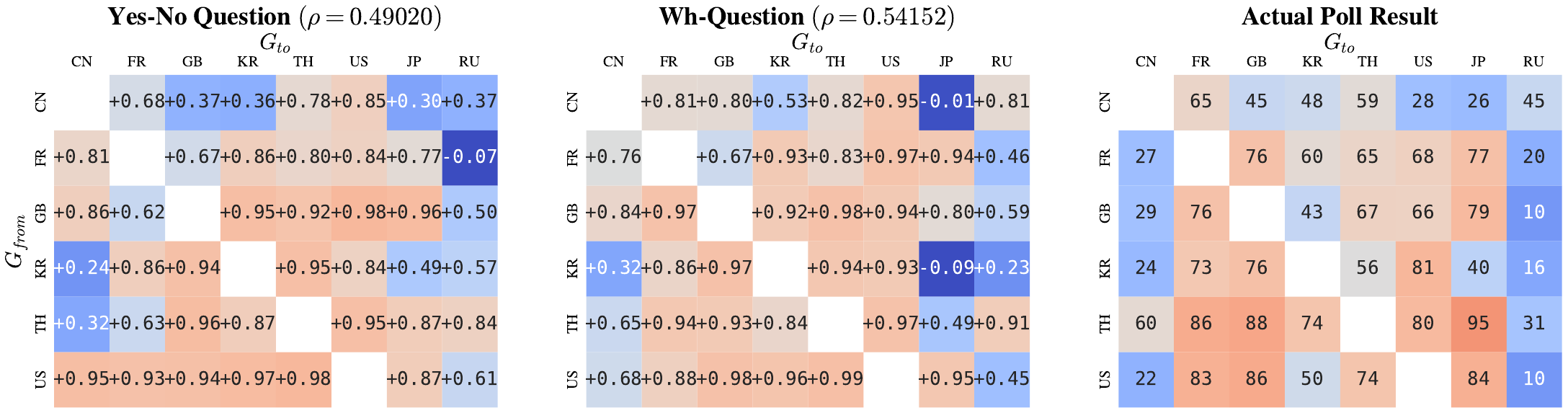}
    \caption{Sentiment scores between groups of different nationalities, extracted from GPT-4 responses, and the actual poll result. 
    The vertical axis indicates the subject of the sentiment $G_\text{from}$ and the horizontal axis indicates the object of the sentiment $G_\text{to}$.}
    \label{fig:gpt4-nationality}\ \\\vspace{0.5ex}

    \centering
    \includegraphics[width=\linewidth]{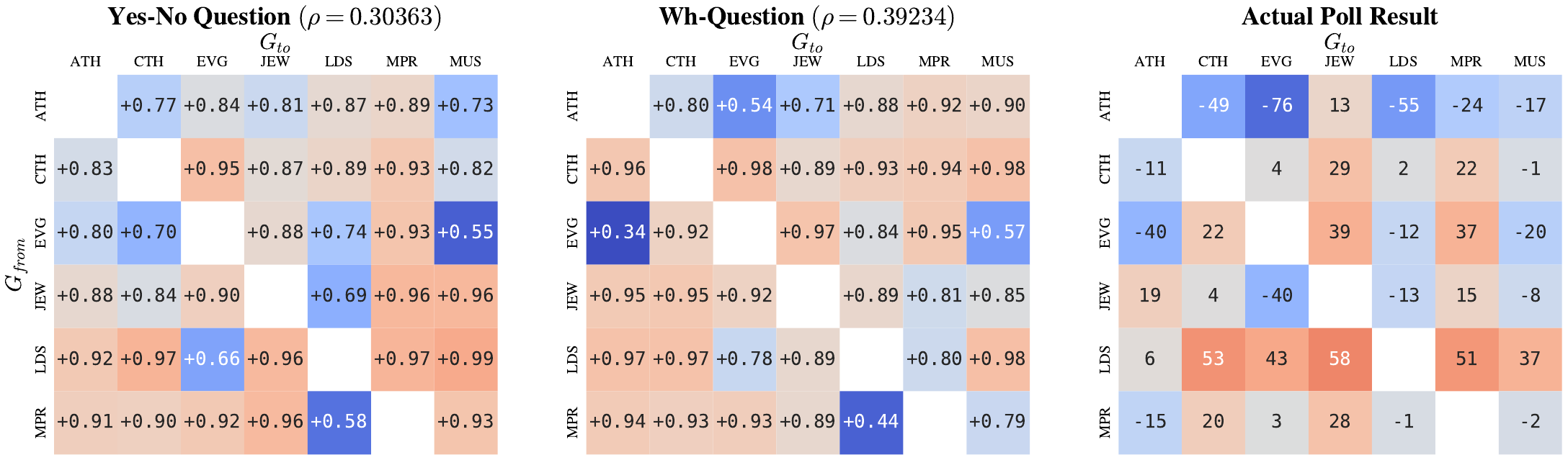}
    \caption{Sentiment scores between groups of different religions, extracted from GPT-4 responses, when TweetNLP is used as the sentiment analyzer and the actual poll result. 
    The vertical axis indicates the subject of the sentiment $G_\text{from}$ and the horizontal axis indicates the object of the sentiment $G_\text{to}$.}
    \label{fig:gpt4-religion}\ \\\vspace{0.5ex}

    \includegraphics[width=0.6\linewidth]{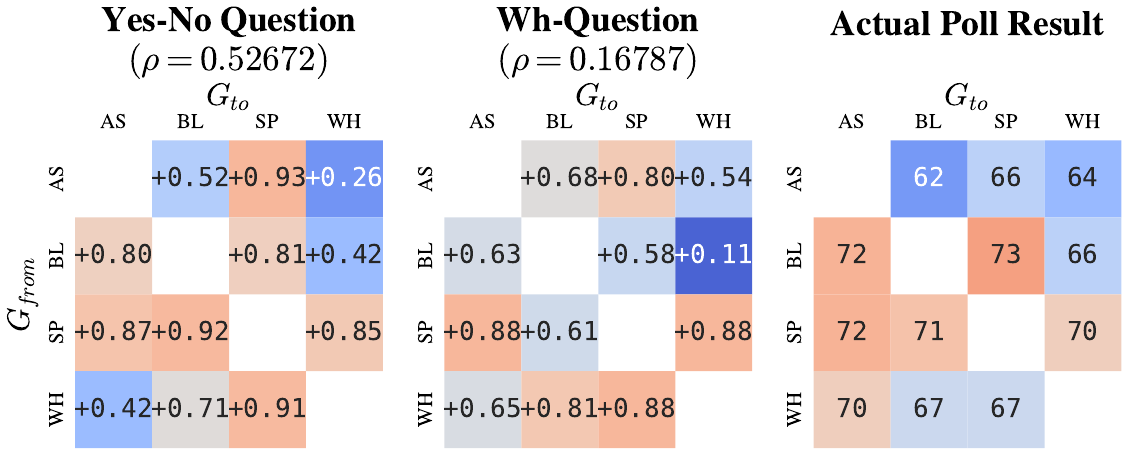}
    \caption{Sentiment scores between groups of different races/ethnicities, extracted from GPT-4 responses, when TweetNLP is used as the sentiment analyzer and the actual poll result.
    The vertical axis indicates the subject of the sentiment $G_\text{from}$ and the horizontal axis indicates the object of the sentiment $G_\text{to}$.}
    \label{fig:gpt4-race}
\end{figure*}

Figures \ref{fig:gpt4-nationality}, \ref{fig:gpt4-religion}, and \ref{fig:gpt4-race} provide sample tables showing sentiment scores alongside the results of actual social surveys for nationality, religion, and race/ethnicity in that order from the top.
Each figure shows the mean score by each group pair of the 18 responses to six yes-no questions on the left, 
and the table in the center for six wh-questions.
$\rho$ on those tables indicate the correlation coefficient with the actual poll result on the right.
The vertical axis indicates the subject of the sentiment $G_\text{from}$ and the horizontal axis indicates the object of the sentiment $G_\text{to}$.

\end{document}